\documentclass[10pt,twocolumn,letterpaper]{article}

\usepackage[pagenumbers]{iccv} % To force page numbers, e.g. for an arXiv version

\newcommand\ve[1]{\mathbf{#1}}
\usepackage{tabularx}
\usepackage{marvosym}
\usepackage{pifont}
\usepackage{multirow}
\usepackage{makecell}

\definecolor{iccvblue}{rgb}{0.21,0.49,0.74}
\usepackage[pagebackref,breaklinks,colorlinks,allcolors=iccvblue]{hyperref}

\def\paperID{***} % *** Enter the Paper ID here
\def\confName{ICCV}
\def\confYear{2025}

\title{A Large-Scale Study on Video Action Dataset Condensation}

\author{
Yang Chen\textsuperscript{1} \quad
Sheng Guo\textsuperscript{2}\quad
Bo Zheng\textsuperscript{2} \quad
Limin Wang\textsuperscript{1, 3, \Letter }\quad\\
$^1$State Key Laboratory for Novel Software Technology, Nanjing University \\
$^2$Ant Group\quad $^3$ Shanghai AI Lab
}
\begin{document}
\maketitle
\newcommand\blfootnote[1]{%	
  \begingroup
  \renewcommand\thefootnote{}\footnote{#1}%
  \addtocounter{footnote}{-1}%
  \endgroup
}

\begin{abstract}
Recently, dataset condensation has made significant progress in the image domain. Unlike images, videos possess an additional temporal dimension, which harbors considerable redundant information, making condensation even more crucial. However, video dataset condensation still remains an underexplored area. We aim to bridge this gap by providing a large-scale study with systematic design and fair comparison. Specifically, our work delves into three key aspects to provide valuable empirical insights: (1) temporal processing of video data, (2) the evaluation protocol for video dataset condensation, and (3) adaptation of condensation algorithms to the space-time domain. From this study, we derive several intriguing observations: (i) labeling methods greatly influence condensation performance, (ii) simple sliding-window sampling is effective for temporal processing, and (iii) dataset distillation methods perform better in challenging scenarios, while sample selection methods excel in easier ones. Furthermore, we propose a unified evaluation protocol for the fair comparison of different condensation algorithms and achieve state-of-the-art results on four widely-used action recognition datasets: HMDB51, UCF101, SSv2 and K400. Our code is available at \url{https://github.com/MCG-NJU/Video-DC}.
\end{abstract}
\blfootnote{\Letter~: Corresponding author (lmwang@nju.edu.cn).}
\section{Introduction}
\label{sec:intro}
\begin{figure}[t]
  \centering
  \includegraphics[width=\linewidth]{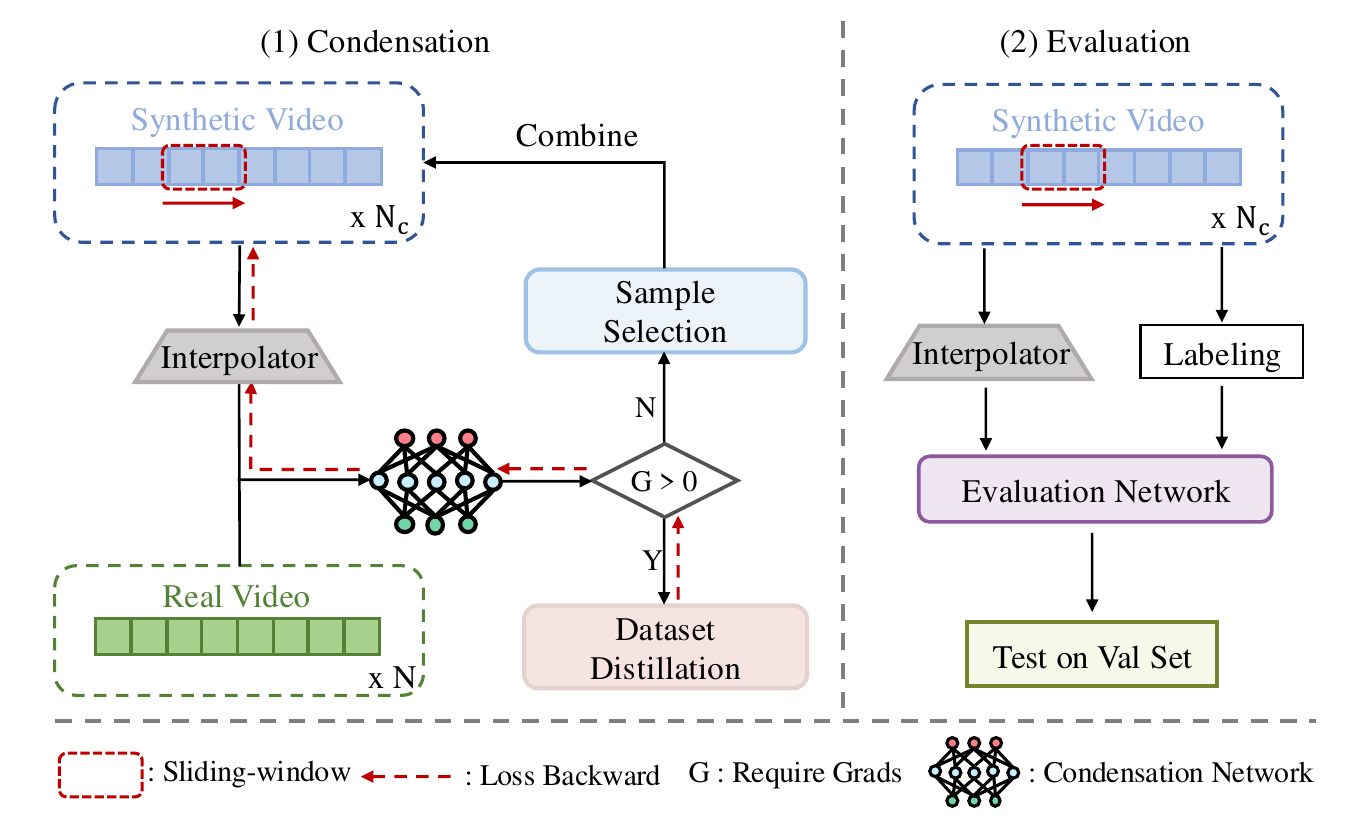}
  \caption{Pipeline of our study. Our study includes three core elements: temporal processing, condensation algorithms and evaluation settings. Temporal processing is the special design for video data, including sampling and interpolation. Real videos comes across a condensation algorithm (categorized into sample selection and dataset distillation, with G indicating the chosen approach) to form the synthetic dataset (left). Then, the synthetic videos combined with the labeling methods train an evaluation network to test on val set as the evaluation metric (right). Both of the phases rely on a neural network.}
  \label{pipeline}
\end{figure}
Dataset condensation, crucial for data-efficient learning, focuses on reducing the size of an original dataset while maintaining the performance of models trained on the condensed version~\cite{DD}. This task addresses challenges associated with training neural networks on large datasets, including high computational costs and storage requirements. Dataset condensation has numerous applications, such as neural architecture search~\cite{NAS}, privacy protection~\cite{privacy}, and more.

Previous research on dataset condensation~\cite{DD,DC,DM,MTT,FRePo,KIP,KIP2,RFAD,DSA,IDC,tesla,CAFE,haba,linba,DATM,RDED,EDC} has primarily focused on image datasets. Some studies have explored extending condensation algorithms to other modalities, such as graph~\cite{graph1, graph2} and text data~\cite{language, language2}.
As a natural extension of images, video serves as a crucial medium for conveying visual information. The additional data in videos increases storage and computational requirements, while also introducing redundant scene and texture information. These factors highlight the increased need for, and the significant potential of video dataset condensation.
% Previous works have applied condensation methods to other domains, such as graph data~\cite{graph1, graph2} and natural language~\cite{language}, while video dataset condensation remains relatively unexplored. 

Wang \etal~\cite{dance} first explore the possibility of applying dataset condensation to video data, parameterizing segmented matching with synthetic frames, real frames, segments and interpolation algorithms. They proposed a two-stage framework that disentangle static and dynamic information. 
% However, they did not take the evaluation protocol into account and the performance of their method is limited. Video dataset condensation remains relatively unexplored.
However, they did not investigate the evaluation settings and primarily conducted experiments on the small-scale HMDB51 and MiniUCF. The performance of video dataset condensation remains limited.

In response to the growing significance of video dataset condensation, we conduct a large-scale empirical study, which involves various temporal processing techniques, diverse condensation algorithms, complex evaluation settings and four action recognition datasets. Based on these, we provide valuable insights and achieve sota results.
% We hope that our systematic study will inspire and guide future research.
% and the performance of our approach significantly outperformed theirs.
% we focus on addressing these problems and also investigate the impact of evaluation settings.
% In response, our work adapts existing algorithms for video data and conducts a large-scale empirical study on video dataset condensation.

~\Cref{pipeline} clearly illustrates the pipeline of our study, which is divided into two main phases: condensation and evaluation. The real dataset is first condensed into a synthetic one by sample selection or dataset distillation methods. Then, the synthetic dataset trains evaluation networks, whose performance on the real validation set serves as the metric to evaluate the condensation algorithms.

Three core elements of the study are also depicted in ~\cref{pipeline}: temporal processing, condensation methods and evaluation settings. For temporal processing, we employ a sampling and interpolation pipeline both on condensation and evaluation phases. For condensation algorithms, we adapt RDED~\cite{RDED}, EDC~\cite{EDC}, DATM~\cite{DATM} to video data, which have demonstrated their effectiveness on image datasets. Representing sample selection methods, RDED directly select samples from real data without optimization. While, the latter two methods, belonging to dataset distillation, synthesize data by optimizing proxy objectives: matching distributions or training trajectories. For evaluation, we utilize action recognition as the downstream task. In addition, we explore various evaluation settings, including labeling methods, data augmentation and loss functions. Based on this, we propose a unified evaluation protocol to enable fair comparison of condensation algorithms.

According to the experiments, we draw intriguing observations: (1) Labeling methods greatly influence final results. Similar conclusions are also drawn in image datasets~\cite{label}. (2) Simple sliding-window sampling is effective for temporal processing. Compared with previous segment sampling~\cite{dance}, our method better guarantee temporal coherence of synthetic videos. 
% (3) Sample selection methods generally outperform dataset distillation methods in most situations while the trend reverses when the condensation ratio is extremely low. This suggests a sub-optimized problem of dataset distillation especially for video data. Once resolved, it holds immense potential for efficient training of large models.
(3) Dataset distillation methods outperform sample selection methods on challenging scenarios, while sample selection methods excel in easier ones. This shows great potential of dataset distillation methods to tackle training efficiency problems, which fits the original intention of dataset condensation.

% In summary, our large-scale study involves the following five aspects:

% (i) Three dataset condensation frameworks (EDC)

% (ii) Sufficient results on four widely-used action recognition datasets;

% (iii) Four architectures for evaluating the generalization of condensed datasets.

% (iv) Ablation experiments on different factors, such as labeling methods, temporal processing, candidates models and evaluation settings;

% (v) State-of-art results of video dataset condensation on HMDB51~\cite{hmdb51}, UCF101~\cite{ucf101}, K400~\cite{kinetics} and SSv2~\cite{ssv2} under the unified evaluation protocol.

In summary, our key contributions are as follows: (1) the establishment of an evaluation protocol for video dataset condensation, (2) the development of temporal processing methods, especially the proposed sliding-window sampling, (3) comprehensive experimental results and in-depth analyses of condensation algorithms, and (4) state-of-the-art performance on widely-used video action datasets: HMDB51~\cite{hmdb51}, UCF101~\cite{ucf101}, SSv2~\cite{ssv2} and K400~\cite{kinetics}.

\section{Related Work}
\label{sec:related work}
\noindent
\textbf{Image Dataset Condensation.} Dataset condensation algorithms can be broadly categorized into sample selection and dataset distillation. Sample selection is further divided into optimization-based and score-based approaches. Optimization-based selection aims to identify a small coreset that effectively captures the diverse characteristics of the full dataset, with methods such as Herding~\cite{herding}, K-center~\cite{kcenter}, Craig~\cite{craig}, and GradMatch~\cite{gradmatch}.
By contrast, score-based selection assigns heuristic measures to instances, like difficulty or the impact on network training. Traditional approaches include Forgetting~\cite{forgetting}, Glister~\cite{glister}, and C-Score~\cite{cscore}. DeepCore~\cite{deepseek} has shown that random selection remains a strong baseline among the above methods. More recently, RDED~\cite{RDED} has achieved remarkable results on various image datasets. It selects samples based on realism and diversity scores, and then concatenates them to form the condensed dataset.

Dataset distillation methods can be categorized into performance matching, distribution matching, and trajectory matching. Due to the inherent bi-level optimization, performance matching methods face scalability challenges~\cite{DD, FRePo, KIP, KIP2, RFAD}, making them less practical for large-scale scenarios. Thus, our focus is primarily on the latter two kinds of methods. Trajectory matching optimizes condensed data by aligning the parameter updates of a network trained on real data and synthetic data. Methods such as DC~\cite{DC}, DSA~\cite{DSA}, MTT~\cite{MTT}, TESLA~\cite{tesla}, and DATM~\cite{DATM} have progressively refined this approach, achieving lossless results on small datasets. Recent concurrent works continue to advance this framework~\cite{MCT, EDF}.
Distribution matching, on the other hand, aligns real and synthetic data across different feature spaces~\cite{DM, CAFE}. SRe2L~\cite{SRe2L} introduced statistical matching to reduce computational and storage costs, and subsequent works such as G-VBSM~\cite{G-VBSM} and EDC~\cite{EDC} further enhanced this approach, leading to improved performance especially on large datasets.

\noindent
\textbf{Condensation on Other Modalities.} Many studies have extended dataset condensation algorithms to other modalities. Graph data, which captures relationships between nodes, has been explored extensively~\cite{graph0,graph1,graph2}. Similarly, text and time series data also draw great interest~\cite{language,language2, time1,time2}. More recently, as a crucial medium for conveying visual information, video is first explored by Wang \etal~\cite{dance}. Building upon this foundation, our work provides insightful results to foster further exploration.

\noindent
\textbf{Action Recognition.} The development of video backbones for action recognition has progressed significantly to better capture spatial and temporal features. C3D~\cite{c3d} introduced 3D convolutions to process both aspects simultaneously but was computationally heavy. I3D~\cite{kinetics} improved efficiency by inflating 2D convolutional filters pre-trained on image data into 3D, benefiting from transfer learning. R(2+1)D~\cite{r2+1d} refined this by separating 3D convolutions into 2D spatial and 1D temporal components, enhancing flexibility and reducing parameters. The SlowFast~\cite{slowfast} network further advanced the field with a dual-pathway design that processes video at different frame rates, capturing both detailed spatial information and fast motion changes. Some transformer-based backbone~\cite{actionclip,timesformer,videovae} have also demonstrated great effectiveness. In this work, we use these video backbones for video dataset condensation.

\section{Method}
% The objective of this work is to study several recent dataset condensation methods to condense video datasets. These methods are originally proposed for image datasets. Generalizing them to space-time, we explore the implementation details and compare them on a common ground to measure their efficacy of dataset condensation. We mainly focus on one sample selection method: RDED[] and two dataset distillation methods: DATM[] and EDC[].
This work investigates recent dataset condensation algorithms for condensing video datasets. For these methods were initially designed for image datasets, we extend them to the space-time domain, examining the implementation details and comparing their performance under consistent conditions to evaluate their efficacy in dataset condensation. Our study primarily focuses on one sample selection method: RDED~\cite{RDED}, and two dataset distillation methods: DATM~\cite{DATM} and EDC~\cite{EDC}.

As in \cref{pipeline}, the process of the dataset condensation can be divided into two main phases. First, the original dataset is condensed into a smaller one using either distillation or selection methods. Next, the condensed dataset is applied to a downstream task to assess the effectiveness of the condensation algorithms. Adapting condensation algorithms from images to videos, we propose sliding-window sampling and interpolation for temporal processing. In this section, we will discuss about the temporal processing and then introduce two phases of dataset condensation separately.

\begin{figure}[t]
    \centering
    \includegraphics[width=\linewidth]{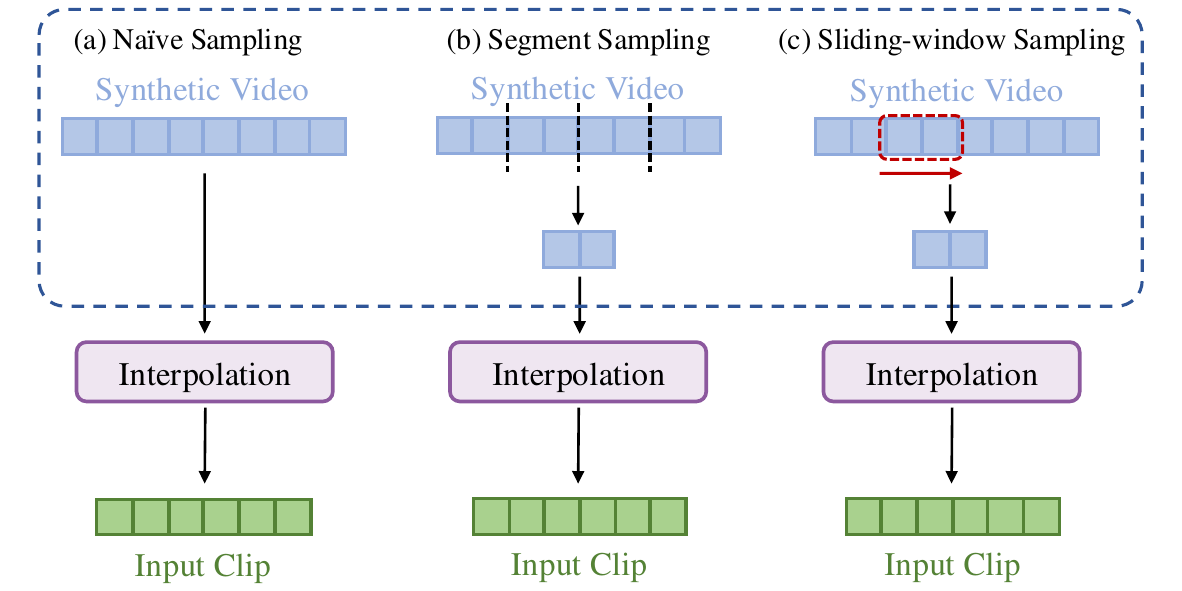}
    \caption{Temporal processing encompasses sampling and interpolation. (a) Naive sampling directly treats the video as the sampled clip. (b) Segment sampling views videos as independent clips. (c) Sliding-window sampling sequentially samples clips along time. Different interpolation methods can then be applied to generate the final input clips.}
    \label{fig:temp}
\end{figure}

\begin{figure*}[t]
  \begin{center}
    \includegraphics[width=0.8\linewidth]{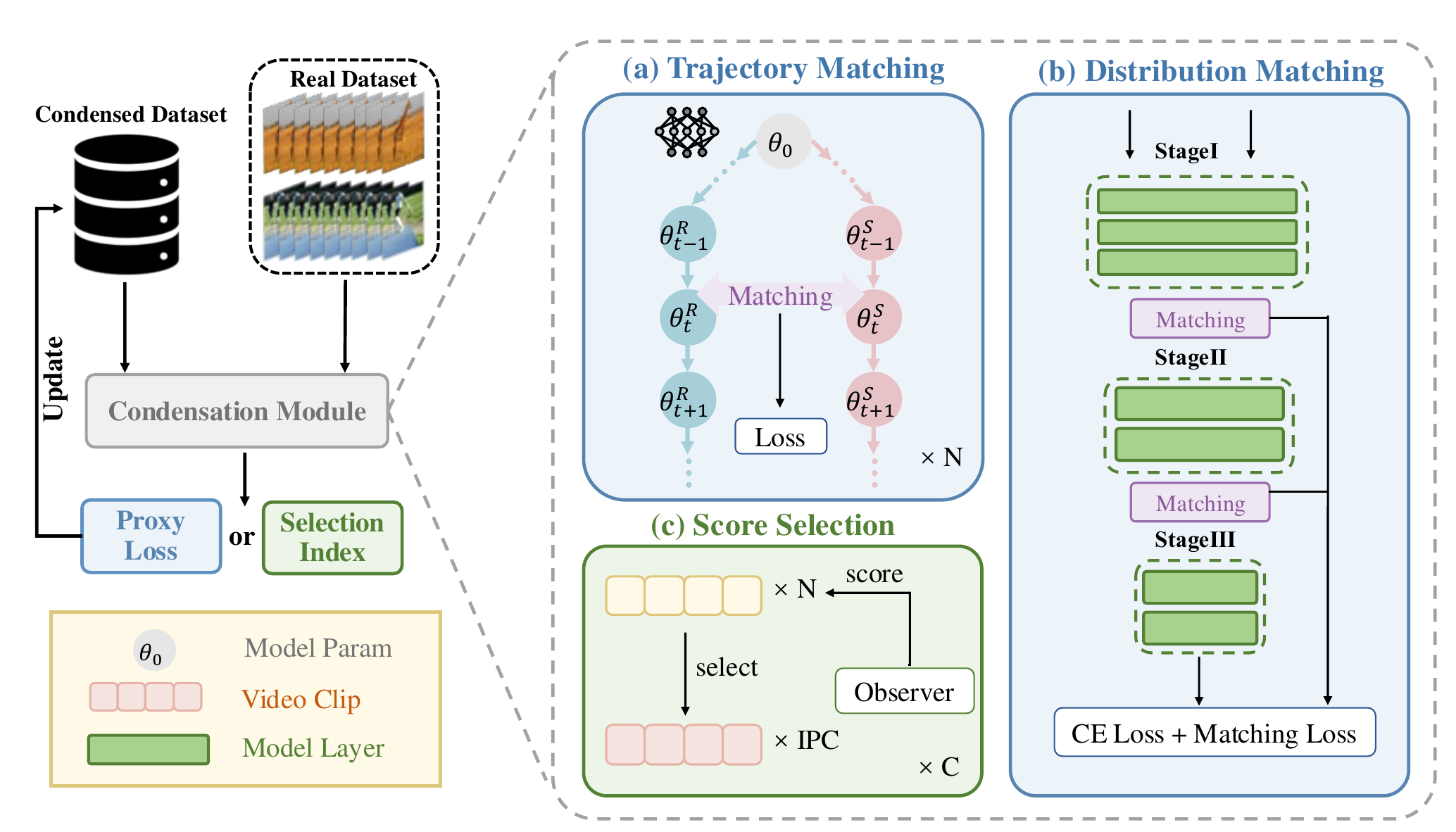}
    \vspace{-10pt}
\end{center}
  \caption{Conceptual visualization of three dataset condensation frameworks applied to video. Trajectory Matching (a) and Distribution Matching (b) in the blue boxes belong to dataset distillation methods, and Score Selection (c) in the green box belongs to sample selection methods. Dataset distillation defines a proxy task to help condense the dataset and uses the output proxy loss (left) to update the condensed dataset, while sample selection directly draws samples from the real dataset to form the condensed one. Trajectory Matching (a) uses a normed L2 loss to describe distance between parameters, Distribution Matching (b) matches the distribution between model layers, and Score Selection (c) scores each sample from the real dataset and concatenates them to form the condensed dataset.}
  \label{fig:main}
\end{figure*}

\subsection{Temporal Processing}
\label{sec:temp}
From images to videos, the processing of temporal dimension is the most crucial topic. As shown in \cref{fig:temp}, temporal processing incorporates sampling and interpolation, getting input clips from synthetic videos. 

For sampling, the most naive method is viewing the whole video as the sampled clip (\cref{fig:temp}a). It can be regarded as the baseline of the sampling method. Wang \etal ~\cite{dance} proposed segment sampling (\cref{fig:temp}b), which splits the synthetic videos as independent segments. This method disrupts the temporal consistency of videos. To overcome this problem, we propose the sliding-window sampling. It sequentially samples video clips along the temporal dimension as shown in ~\cref{fig:temp}c. Thus, the co-optimization of overlapping clips ensures the consistency and coherence of the synthetic videos.

Additionally, the interpolation methods encompass none, duplication, linear interpolation and even learnable interpolation. It aligns the length of the sampled clips with the input clips and better condenses the temporal information of synthetic videos.

The two processes help condense the temporal information of videos, which is essential for video dataset condensation. Let $N$ and $T_m$ denote the number of videos and the average number of frames in the real dataset, respectively, and $N_c$ and $T_c$ represent those values for the condensed dataset. Thus, $\frac{N_c}{N}$ represents the instance compression ratio, $\frac{T_c}{T_m}$ represents the temporal compression ratio, and the total condensation ratio is given by $\frac{N_c \times T_c}{N \times T_m}$. We will further analyze the trade-off between $N_c$ and $T_c$ in ~\cref{exp: temp}.
% From images to videos, the processing of the temporal dimension is the most crucial topic. Like in action recognition, the clip input needs to be sampled and augmented before being passed into the network. To enhance temporal compression, an interpolation module can be added after sampling to adjust the compressed video to the required number of frames~\cite{dance}. Wang \etal ~\cite{dance} proposed splitting the compressed video into separate segments, with each segment serving as a training video clip. By contrast, we propose using a slide-window sampling across the synthetic video, enabling more efficient use of the condensed data. 

% Previous dataset condensation methods mainly focus on condensing between instances. However, in the case of videos, there is often much redundant information along the temporal dimension, making it crucial to explore temporal compression. Maintaining the same spatial resolution, we condensed the dataset to $N_c$ samples, each with $T_c$ frames. Then, the total condensation ratio can be expressed as $\frac{N_c \times T_c}{N \times T_m}$, where N is the number of videos in the original dataset, and $T_m$ is the average number of frames per original video. We further define $\frac{N_c}{N}$ as the instance compression ratio and $\frac{T_c}{T_m}$ separately as temporal compression ratio. These two factors are ablated in \cref{tab:IPC vs T} and \cref{tab:pureT} explores the effectiveness of pure temporal compression.

\subsection{Condensation Algorithms}
% Dataset condensation can be divided into two main kinds: sample selection and dataset distillation. The key difference is whether to optimize the condensed data. In this part, we mainly introduce sample selection algorithms RDED[], dataset distillation algorithms EDC[], DATM[] and relabel strategies used in them.
Given a large, real dataset $\mathcal{D}^\mathcal{T} := \{\ve{x}^\mathcal{T}_i,\ve{y}^\mathcal{T}_i\}_{i=1}^{\lvert \mathcal{D}^\mathcal{T} \rvert}$ consisting of images $\mathcal{X}^\mathcal{T}$ and labels $\mathcal{Y}^\mathcal{T}$, dataset condensation aims to synthesize a smaller dataset $\mathcal{D}^\mathcal{S} := \{\ve{x}^\mathcal{S}_i, \ve{y}^\mathcal{S}_i\}_{i=1}^{\lvert \mathcal{D}^\mathcal{S} \rvert}$, which includes images $\mathcal{X}^\mathcal{S}$ and labels $\mathcal{Y}^\mathcal{S}$, such that a model trained on $\mathcal{D}^{\mathcal{S}}$ exhibits comparable generalization performance to one trained on $\mathcal{D}^{\mathcal{T}}$. The objective of dataset condensation can be formally expressed as:
\begin{equation}
    \begin{aligned}
        &\mathcal{D}^{\mathcal{S}} = \mathop{\arg\min}_{\mathcal{D}^{\mathcal{S}}} \mathbb{E}_{(\ve{x}^{\mathcal{T}},\ve{y}^{\mathcal{T}}) \sim \mathcal{D}^{\mathcal{T}}} [\ell(\phi_{\mathbf{\theta}^{\ast}}(\ve{x}^{\mathcal{T}}), \ve{y}^{\mathcal{T}})] \\
        &s.t.\ \mathbf{\theta}^{\ast} = \mathop{\arg\min}_{\mathbf{\theta}}\mathbb{E}_{(\ve{x}^{\mathcal{S}},\ve{y}^{\mathcal{S}}) \sim \mathcal{D}^{\mathcal{S}}}[\ell(\phi_{\mathbf{\theta}}(\ve{x}^{\mathcal{S}}), \ve{y}^{\mathcal{S}})],
    \end{aligned}
    \label{eq:dc}
\end{equation}
% \[
% \begin{aligned}
% &\mathcal{D}_{syn} = \mathop{\arg\min}_{\hat{X}, \hat{Y}} \mathbb{E}_{(x, y) \sim \mathcal{D}{real}} \ell(\phi_{\mathbf{\theta}^\ast}(x), y) \\
% &\text{subject to} \quad \mathbf{\theta}^\ast = \mathop{\arg\min}_{\mathbf{\theta}} \ell(\phi_{\mathbf{\theta}}(\hat{x}), \hat{y}),
% \end{aligned}
% \]
where $\ell$ denotes the loss function for the specific task, $\phi_{\mathbf{\theta}}$ represents the model, and $\mathbf{\theta}^\ast$ denotes the optimal parameters after training on $\mathcal{D}^{\mathcal{S}}$. This objective is challenging to optimize directly. 
Consequently, previous works introduce proxy tasks to optimize condensed datasets. Specifically, dataset distillation methods use the proxy loss $\mathcal{L}$ to describe differences between real and synthetic data, while sample selection methods typically design heuristic objectives for optimization. We then present the proxy tasks of various condensation algorithms.
% We decompose the synthesis of videos and labels, discussing each aspect separately. As summarized in \cref{sec:related work}, condensation algorithms can be classified into sample selection and dataset distillation. Additionally, we categorize the relabel strategies they use into three main types: hard label, soft label and multiple soft label. In this subsection, we will discuss about these methods carefully. \Cref{fig:main} visualizes the three types of condensation algorithms.

\noindent
{\bf DATM}~\cite{DATM} (\cref{fig:main}a) is a representative of matching training trajectories. The proxy task is to match the training trajectories of condensation models trained on $\mathcal{D}^{\mathcal{T}}$ and $\mathcal{D}^{\mathcal{S}}$. Specifically, let $\{ \theta^{R}_{t} \}_{0}^{n}$ denote the training trajectory obtained by $\mathcal{D}^{\mathcal{T}}$, where $\theta_t^{R}$ is the parameters of the condensation network at training step t. Similarly, $\theta_t^{S}$ denotes the parameters trained on $\mathcal{D}^{\mathcal{S}}$ at time step t. 

In each iteration of the distillation, sampling time step t from 0 to n, the proxy loss is:
\begin{equation}
    \mathcal{L}=\frac{\|{\theta}^S_{t+N}-\theta_{t+M}^R\|_2^2}{\|\theta_t^R-\theta_{t+M}^R\|_2^2},
    \label{eq:datm}
\end{equation}

\noindent
where M and N is the hyper-parameter of optimization step on $\mathcal{D}^{\mathcal{T}}$ and $\mathcal{D}^{\mathcal{S}}$ for the condensation network.

Specifically, to better align the difficulty of distillation as the process progresses, DATM gradually increases the time steps of the initial parameters, ensuring smoother and more effective optimization.

% [using equations]
\noindent
\textbf{EDC}~\cite{EDC} (\cref{fig:main}b) follows SRe2L~\cite{SRe2L} to generalize distribution matching to statistical matching. The proxy task is to matching the statistics of original and condensed datasets. 
\begin{equation}
% \footnotesize
\begin{aligned}
    \mathcal{L}= &||p(\mu|\mathcal{X}^\mathcal{S})-p(\mu|\mathcal{X}^\mathcal{T})||_2 \\
    &+||p(\sigma^2|\mathcal{X}^\mathcal{S})-p(\sigma^2|\mathcal{X}^\mathcal{T})||_2, \ s.t.\ \mathcal{L}\sim \mathbb{S}_\textrm{match}, 
    % &\quad \mathcal{X}^{\mathcal{S}*} = \operatorname*{arg\,min}_{\mathcal{X}^{\mathcal{S}}} \mathbb{E}_{\mathcal{L}\sim \mathbb{S}_\textrm{match}}[\mathcal{L}(\mathcal{X}^{\mathcal{S}},\mathcal{X}^{\mathcal{T}})] ,\\
\end{aligned}
\end{equation} 
% \begin{multline}
%     \mathcal{L}= ||p(\mu|\mathcal{X}^\mathcal{S})-p(\mu|\mathcal{X}^\mathcal{T})||_2 \\
%     +||p(\sigma^2|\mathcal{X}^\mathcal{S})-p(\sigma^2|\mathcal{X}^\mathcal{T})||_2, \ s.t.\ \mathcal{L}\sim \mathbb{S}_\textrm{match}, 
%     % &\quad \mathcal{X}^{\mathcal{S}*} = \operatorname*{arg\,min}_{\mathcal{X}^{\mathcal{S}}} \mathbb{E}_{\mathcal{L}\sim \mathbb{S}_\textrm{match}}[\mathcal{L}(\mathcal{X}^{\mathcal{S}},\mathcal{X}^{\mathcal{T}})] ,\\
% \end{multline}
where $\mathbb{S}_\textrm{match}$ denotes the extensive collection of statistical matching operators, which operate across a variety of network architectures and layers as described by G-VBSM~\cite{G-VBSM} and EDC~\cite{EDC}. Here, $\mu$ and $\sigma^2$ are defined as the mean and variance, respectively.
% The core concept of distribution matching is to align data distribution of $\mathcal{D}_{real}$ and $\mathcal{D}_{syn}$ in different feature space of condensation models, ensuring that synthetic data achieves optimization effects comparable to real data. For scalability, EDC and its predecessors~\cite{SRe2L, G-VBSM} simplify the distribution matching process by focusing on matching the mean and variance of features. By pre-computing the mean and variance of features across different modules from the condensation model, the distillation process becomes extremely efficient and computationally lightweight. Due to its efficiency, we select EDC as the representative distribution matching algorithm. More details about EDC can be found in the Appendix.

\noindent
\textbf{RDED}~\cite{RDED} (\cref{fig:main}c) emphasizes the realism and diversity of the condensed dataset. It selects the most realistic instances from a set of samples with a greedy algorithm and concatenates them to form the condensed dataset. These two designs ensure the realism and diversity respectively.
\begin{equation}
    \begin{aligned}
        &\{\ve{x}_k^i\}_{k=1}^{N} = \mathop{\arg\max}_{\ve{x}_k}[\sum_{k=1}^{N}\mathcal{F}(\ve{x}_k)], \quad \ve{x}_k \in \mathcal{X}^{\mathcal{T}}_i, \\
        & \mathcal{X}^{\mathcal{S}} = \bigcup_{i=1}^\mathbf{C}\mathcal{X}^\mathcal{S}_i, \quad \mathcal{X}^{\mathcal{S}}_i = \{\ve{x}_j^i = \text{concat}(\{\ve{x}_k^i\}_{k=1}^{N})\}_{j=1}^{\text{IPC}},
    \end{aligned}
    \label{eq:rded}
\end{equation}
where $\mathcal{F}$ is the score function measuring the realism of instances, $\mathbf{C}$ denotes the number of classes and concat($\cdot$) concatenates N samples together. 
% The training-free approach also leads to high efficiency.
% The realism of a clip $\xi_{i,k}$ is defined as $-\ell(\phi_{\boldsymbol{\theta}_{\mathcal{T}}}(\xi_{i,k}), y_i)$, where $\phi_{\boldsymbol{\theta}_{\mathcal{T}}}$ represents the pretrained model and $y_i$ is the human-annotated label.

Previous studies~\cite{deepseek} show that many carefully designed sample selection methods do not have obvious advantages over random selection. However, RDED achieves remarkable results with this simple design, and the training-free method also guarantees high efficiency.
% Thus, we choose RDED as the representative sample selection algorithm.

\noindent
\textbf{Implementation Specifics.} We draw an analogy between clips in videos and cropped images. For dataset distillation methods, we update the batch of sampled clips in each iteration. We extend the spatial statistics in EDC to the space-time domain. By contrast, for concatenation in RDED, we only concatenate video tubes along the spatial dimension. More details can be found in ~\cref{app:train}.

\begin{table}[t]
  \centering
  \begin{tabular}{ccccc}
    \toprule
    Dataset & \#videos & $n_{mean}$ & $T_{mean}$ & $T_{median}$ \\
    \midrule
    HMDB51 & 3,570 & 70 & 97 & 79 \\
    UCF101 & 9,537 & 94 & 187 & 168 \\
    SSv2 & 168,913 & 970 & 45 & 44 \\
    K400 & 240,436 & 601 & 286 & 300 \\
    \bottomrule
  \end{tabular}
  \caption{Statistics of action recognition video datasets. $n_{mean}$ represents the average number of videos per class. $T_{mean}$ and $T_{median}$ are the mean and median of frames per video.}
  \label{dataset statistics}
\end{table}

\begin{figure}
    \vspace{-10pt}
    \centering
    \includegraphics[width=\linewidth]{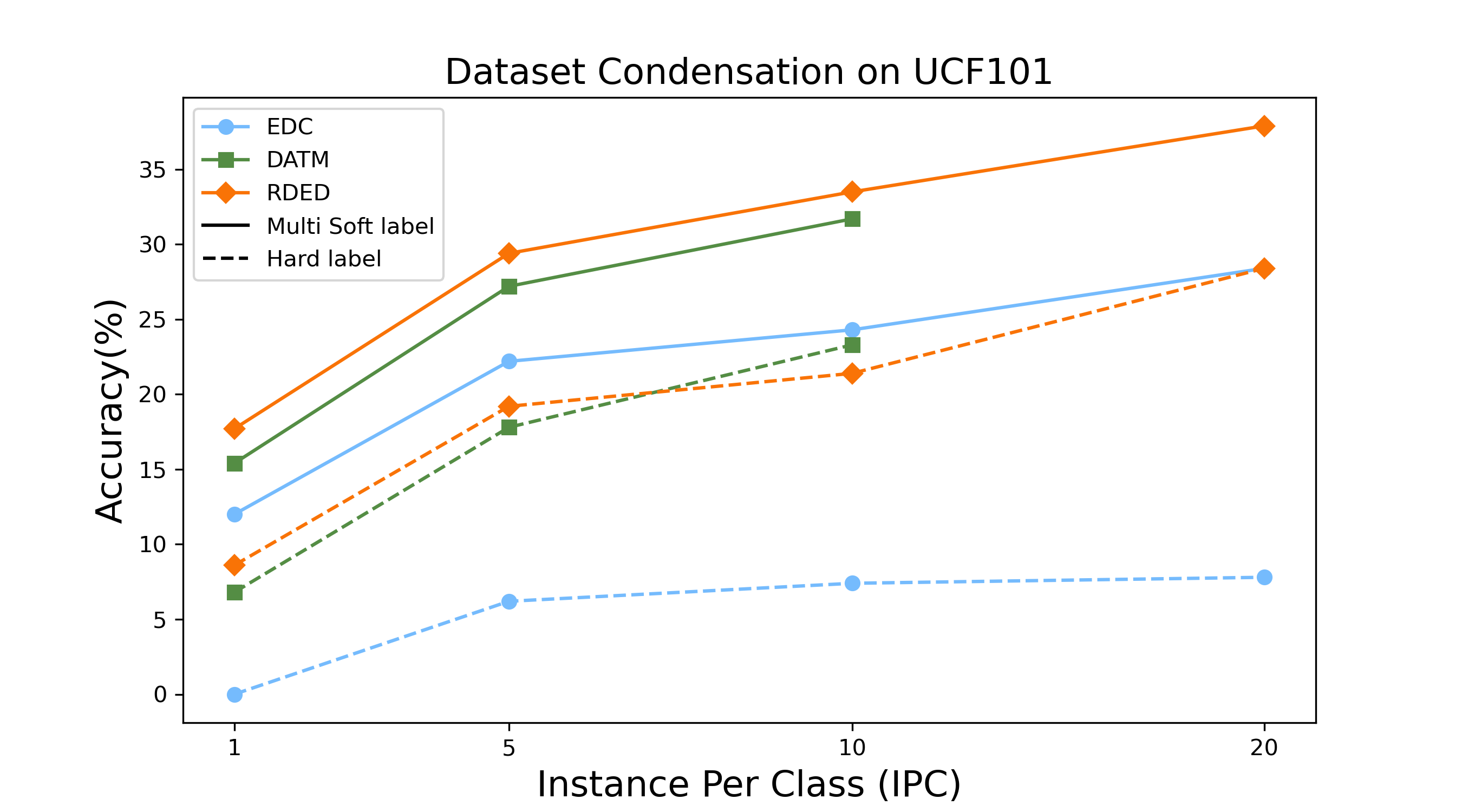}
    \caption{Performance curve of various condensation algorithms with different labeling methods on UCF101, illustrating the comparison among condensation algorithms and highlighting the significant impact of labeling methods.}
    \label{fig:relabel}
\end{figure}

\subsection{Evaluation Protocol}
% use equation to formulate

% This work mainly focuses on the video dataset condensation of action recognition. The evaluation process of dataset condensation includes training a brand new neural network for action recognition. This process is more complicated than normal one, and different training settings greatly influence the final metric. Dataset condensation aims to save the computation and storage costs of training at the same time. Considering the objective, we set the number of iterations around 20\% of that of the original dataset. Under this setting, the neural network at evaluation process may not saturate, but the metric is fair and can reflect the performance under real-world downstream tasks.

This work primarily focuses on video dataset condensation for action recognition. The evaluation process involves training a new neural network specifically for action recognition, making it more complex than other tasks. Consequently, different training settings significantly impact the final metric. 
Thus, we conduct an experimental analysis of the factors that have the most significant impact on the results, including labeling methods, data augmentation, and loss functions. According to the results, we propose a unified evaluation protocol for fair comparison of different condensation algorithms.

\noindent
\textbf{Labeling methods} include hard labeling, soft labeling and multi soft labeling (Multi-SL). \underline{Hard labeling}~\cite{DD} assigns each video a one-hot encoded label that remains unchanged during the dataset condensation process. This method is the most straightforward and intuitive, serving as the fundamental method. \underline{Soft labeling}~\cite{tesla, DATM} utilizes logits as labels to enhance performance. 
The logits can be be obtained from a pre-trained teacher model or defined as learnable vectors optimized alongside the condensed videos.
Finally, \underline{Multi-SL}~\cite{SRe2L, G-VBSM, EDC} adopts the philosophy of knowledge distillation. For the same video, multiple inputs are generated through sampling and data augmentation, each assigned different logits as labels. 
By doing so, it captures diverse aspects of the video under various augmented views, enhancing the model’s generalization ability.

% Dataset condensation not only aims to save storage consumption but also the training time when utilizing condensed datasets. To achieve this, we limit the number of iterations to approximately 20\% of that used for the original dataset. Under these conditions, the neural network may not fully converge, but the metric remains fair and can better reflect performance in real employment situation.

Additionally, data augmentation and loss functions significantly impact the results. We conduct ablation studies on them with different condensation and labeling methods to broadly demonstrate that: (1) comparisons between condensation algorithms are consistent across different evaluation settings, (2) evaluation settings substantially affect the final metrics, and (3) employing different evaluation settings when comparing condensation algorithms may lead to incorrect conclusions. Therefore, we propose a unified evaluation protocol based on our experimental results. Details of the evaluation protocol can be found in ~\cref{app:eval}.

% Moreover, we also propose a new metric which can better reflect the generalization of the condensed dataset. Previous works mainly evaluate the condensed dataset by training the neural network used at the condensing period. We use the condensed dataset to fine-tune an unseen architecture. This metric can reflect the generalization to different architectures and downstream tasks.

\section{Experiment}
Our work builds upon numerous existing methods for image dataset condensation, so we retain their original names for simplicity. In this section, unless otherwise specified, all methods refer to their video-version implementations.

\noindent
{\bf Datasets.} We conduct experiments on HMDB51~\cite{hmdb51}, UCF101~\cite{ucf101}, SSv2~\cite{ssv2} and K400~\cite{kinetics}. 
\Cref{dataset statistics} shows the statistics of the action recognition datasets. HMDB51 consists of 51 action categories, with 70 videos per category. UCF101 contains 101 categories, with an average of 94 videos per category. SSv2 consists of 174 categories, with an average of up to 970 videos per category. K400 consists of 400 categories, with an average of 600 videos per category. With a suitable scale on video duration and number, UCF101 is the main dataset for ablation. 
Given the current progress in video dataset distillation, SSv2 and K400 are sufficient for validating the scalability of dataset condensation algorithms. Additionally, SSv2 emphasizes temporal dynamics and motion patterns, making it particularly useful for evaluating the condensation of motion information.

\noindent
{\bf Experiment Details.} 
For training, we mainly use a four-layer MiniC3D following the implementation in ~\cite{dance}. Other model architectures are explored in ~\cref{tab: architecture}. For evaluation, we report the main results using MiniC3D, and cross-architecture generalization results with R(2+1)D, I3D and SlowOnly. The latter three models use the ResNet-18 backbone.
For HMDB51 and UCF101, we crop and resize the frame to 112 $\times$ 112. For SSv2 and K400, the frames are cropped and resized to 56 $\times$ 56, considering memory and computation consumption. The default data augmentation we use is resized crop and horizontal flip.
\begin{table}[t]
  \centering
  \begin{tabular}{c|cc|cc|cc}
    \toprule[1pt]
     & \multicolumn{2}{c|}{Hard Label} & \multicolumn{2}{c|}{Soft Label} & \multicolumn{2}{c}{Multi-SL} \\
    CutMix & w/ & w/o & w/ & w/o & w/ & w/o \\
    % \cmidrule(lr){1-7}
    \hline
    EDC\textsuperscript{\textdagger} & 4.5 & \underline{4.6} & 3.0 & \underline{6.0} & 5.7 & \underline{7.0} \\
    DATM & 6.7 & \underline{6.8} & 9.7 & \underline{12.1} & 14.9 & \underline{15.4} \\
    RDED & \underline{10.2} & 9.7 & 11.0 & \underline{14.3} & 16.1 & \textbf{\underline{17.7}} \\
    \bottomrule[1pt]
  \end{tabular}
  \caption{Ablation results of CutMix augmentation on UCF101. Not using CutMix consistently improve the accuracy with only one exception (RDED hard label). EDC\textsuperscript{\textdagger} represents a simple version of EDC without category-wise matching. }
  \label{tab:CutMix}
\end{table}

\begin{table}[t]
  \centering
  \begin{tabular}{c|cc|cc}
    \toprule[1pt]
     & \multicolumn{2}{c|}{Soft Label} & \multicolumn{2}{c}{Multi-SL} \\
    Loss & KL & MSE-GT & KL & MSE-GT \\
    % \cmidrule(lr){1-7}
    \hline
    EDC\textsuperscript{\textdagger} & 4.8 & \underline{6.0} & 5.5 & \underline{7.9} \\
    DATM & 11.3 & \underline{13.7} & 14.7 & \underline{15.4} \\
    RDED & \underline{15.0} & 14.3 & \textbf{\underline{18.2}} & 17.7 \\
    \bottomrule[1pt]
  \end{tabular}
  \caption{Ablation results of loss function on UCF101. MSE-GT loss generally outperforms KL loss, except RDED. For not affecting the comparison between method, we choose MSE-GT loss for simplification. EDC\textsuperscript{\textdagger} is a simple version of EDC without category-wise matching.}
  \vspace{-8pt}
  \label{tab:Loss func}
\end{table}

\subsection{Effect of Evaluation Settings}
In this section, we analyze key aspects of the evaluation process, including labeling methods, data augmentation, and loss functions. Our experiments highlight the importance of maintaining a consistent evaluation setting, leading to the proposal of a unified evaluation protocol. 
% \TODO{All specifics of the evaluation protocol can be found in the Appendix.}

\noindent
{\bf Labeling Methods.} ~\Cref{fig:relabel} shows the performance curve of various condensation algorithms with different labeling methods. It can be vividly seen that labeling methods influence performance a lot, even greater than condensation algorithms (EDC with Multi-SL outperforms DATM with hard labeling). We provide more results with different labeling methods on ~\cref{tab:CutMix} and ~\cref{tab:Loss func}, further confirming the impact of labeling methods. In terms of labeling methods, Multi-SL performs the best, followed by soft labeling, and then hard labeling. It is unacceptable that an evaluation setting outweighs the algorithm itself. Thus, we emphasize the importance of a unified evaluation protocol. We consistently use Multi-SL in our ablations, and soft labeling in ~\cref{tab:sota} for fair comparison with the previous work~\cite{dance}.

% \begin{table}[t]
%     \centering
%     \begin{tabular}{cccccc}
%         \toprule[1pt]
%             $T_c$ & 4 & 8 & 16 & 24 & All \\
%             \hline
%             HMDB51 & 17.3 & 23.3 & \textbf{24.0} & 23.1 & 26.1 \\
%             UCF101 & 33.5 & \textbf{43.7} & 42.8 & 42.3 & 52.6 \\
%             SSv2 & 27.0 & \textbf{40.4} & 39.1 & 38.4 & 47.5 \\
%             K400 & & & & & \\
%             \bottomrule[1pt]
%             % \hline
%     \end{tabular}
%     \caption{Pure Temporal Compression (${N_c=N}$) based on RDED. All denotes the results on the full dataset.}
%     \label{tab:pureT}
% \end{table}

\noindent
{\bf Augmentation and Loss Function.} \Cref{tab:CutMix} presents an ablation study on the use of CutMix~\cite{cutmix}, a strong data augmentation applied in previous works~\cite{G-VBSM, EDC, RDED}. \Cref{tab:Loss func} ablates two loss functions widely used in previous dataset condensation works~\cite{G-VBSM, DATM, RDED, EDC}: KL-Divergence and MSE-GT. All of the experiments are done in the setting of IPC=1 on UCF101. We show the results of three condensation algorithms with different labeling methods and make the following observations:
\begin{table}[t]
    \centering
    \setlength{\tabcolsep}{4pt}{
    \begin{tabular}{lc|ccc}
    \toprule[1pt]
       Sampling  & Interopolation & RDED & EDC & DATM\\
         \hline 
        Naive & \ding{55} & 11.2 & 10.0 & 7.2 \\
        Segment & \ding{55} & 10.6 & 10.9 & 7.3 \\
        Sliding-window & \ding{55} & \textbf{11.5} & \textbf{12.0} & \textbf{8.5}\\
        Sliding-window & \checkmark & 9.0 & 11.8 & 5.5\\
        \bottomrule[1pt]
    \end{tabular}}
    \caption{Ablation results of temporal processing on K400. Sliding-window sampling outperforms segment sampling and brings larger improvements on synthetic data. Linear interpolation does not work for all methods.}
    \label{tab:inter}
\end{table}

\begin{table}[t]
    \centering
    \begin{tabular}{cccc}
            \toprule[1pt]
            Interpolation & None & Duplication & Linear \\
            \hline
            EDC & \textbf{11.2} & 8.9 & 9.3 \\
            DATM & \textbf{15.3} & 10.5 & 9.0 \\
            RDED & \textbf{17.5} & 12.3 & 15.2 \\
            \bottomrule[1pt]
    \end{tabular}
    \caption{Ablation results of interpolation algorithms on SSv2. It shows that training-free interpolators are ineffective.}
    \label{tab:duplinear}
\end{table}

(1) Using MSE-GT loss without CutMix augmentation generally leads to better performance. Through all the comparison groups, labeling methods contribute up to an 8.6\% improvement, followed by 3.3\% from augmentation and 2.4\% from loss functions, in that order.
% can bring at most 8.6\% improvement, 3.3\% for augmentation and 2.4\% for loss functions in sequence.

(2) Whatever the evaluation setting, RDED performs better than DATM, and EDC\textsuperscript{\textdagger} performs worst. That is to say the comparison between methods is consistent under various unified evaluation settings.

(3) Although RDED generally performs best under the same setting, DATM without CutMix provides 12.1\% accuracy surpassing the performance of RDED with CutMix (11.0\%) under soft labels. There remains such examples. This shows that evaluation settings can affect the conclusion of comparison.

To conclude, evaluation settings for dataset condensation makes a great influence to the final metric, sometimes even larger than the condensation algorithm itself. Thus, it is urgent to establish an evaluation protocol for video dataset condensation. Following our experiments, we adopted a evaluation protocol with Multi-SL labeling, without CutMix, and using MES-GT loss. Further details of the evaluation protocol can be found in \cref{app:eval}.
% To achieve the goal of reducing training time, we adjust the batch size to make sure the number of iterations is around 20\% of that when trained on the original dataset.
% To reduce training time, we adjust the batch size to ensure the number of iterations is around 20\% of that required when training on the original dataset.

\subsection{Effect of Temporal Processing}
\label{exp: temp}
Previous experiments employ naive temporal processing, where synthetic videos are directly used as model inputs. In this subsection, we investigate advanced temporal processing methods, including sampling and interpolation (\cref{sec:temp}), and present ablation results on instance compression and temporal compression.

\Cref{tab:inter} first ablates different temporal processing methods on K400, and we also provide results on SSv2 in ~\cref{app:exp}. The first row is the baseline setting that we used in previous subsection. The second and the third rows apply different sampling methods as introduced in ~\cref{sec:temp}. The fourth row adds linear interpolation based on the third row. From the results, we conclude that: 
(1) Our proposed sliding-window brings consistent improvements and outperforms segment sampling. This aligns with our analysis in ~\cref{sec:temp}. 
(2) Linear interpolation brings performance drop. It appears that training-free interpolation is ineffective for video dataset condensation. 

~\Cref{tab:duplinear} further verifies the second conclusion above. It provides results of three algorithms with different interpolation algorithms. We observe that not using interpolation yields the best performance among these training-free interpolators. 
For trainable interpolators, we encourage further exploration of their performance.
% Thus, we encourage further efforts to explore the performance of trainable interpolator as in ~\cite{dance}. 
Following our experiments, we only employ interpolation when the number of video frames $T_c$ is lower than the model's input length 8 in subsequent studies.
\begin{table}[t]
    \centering
    \begin{tabular}{ccccc}
            \toprule[1pt]
            [IPC, $T_c$] & [8, 4] & [4, 8] & [2, 16]  & [1, 32]  \\
            \hline
            HMDB51 & 15.2 & \textbf{18.3} & 17.5 & 17.8 \\
            UCF101 & 22.9 & \textbf{26.1} & 24.3 & 23.3  \\
            SSv2 & 23.5 & \textbf{25.3} & 22.6 & 19.5 \\
            K400 & 14.3 & \textbf{16.2} & 15.4 & 14.1 \\
            \bottomrule[1pt]
    \end{tabular}
    \caption{Trade-off between IPC and $T_c$ based on RDED. Generally, IPC outweighs $T_c$, and [4,8] performs best on all the datasets. When $T_c$ is set to four, the drop in performance is due to the use of additional interpolation. Note that IPC $\propto N_c$.}
    \vspace{-6pt}
    \label{tab:IPC vs T}
\end{table}

Then, \cref{tab:IPC vs T} shows the trade-off between IPC and $T_c$ based on RDED. The training iteration of the evaluation model and the total frame number remain consistent across different results. Additional interpolation is applied to [8,4], matching the frame number to the model's input length. Generally, larger IPC brings better results except [8,4]. Considering the interpolation brings sharp performance drop by previous analysis, we conclude that the number of videos is more important than the number of frames.
% we conclude that IPC generally outweighs the number of video frames $T_c$. In another word, instance compression is more important than temporal compression.

% Finally, we also provide results of RDED maintaining the number of videos in ~\cref{tab:pureT}. It offers a complementary perspective on the importance of temporal compression.

\subsection{Ablation of Condensation Models}
In the dataset condensation pipeline, neural networks play two key roles: aiding the condensation process and being trained on the condensed dataset to evaluate the effectiveness of the condensation algorithm. Previous experiments use MiniC3D for both roles. However, in dataset distillation, using multiple condensation models often enhances performance. This subsection investigates the impact of different condensation models on dataset distillation performance on UCF101.

We try MiniC3D, R(2+1)D, I3D and SlowOnly as condensation models on ~\cref{tab: architecture}. The results show that I3D and SlowOnly boost the performance when combined with MiniC3D, but R(2+1)D has a negative influence. This may be due to R(2+1)D explicitly separates spatial and temporal modeling, creating a significant structural difference from MiniC3D. 
As a result, it is challenging to jointly optimize condensed data using both architectures simultaneously. Furthermore, the final row demonstrates that the improvements achieved through the integration of I3D and SlowOnly can be accumulated.
\begin{table}[t]
    \centering
    \setlength{\tabcolsep}{3pt}{
    \begin{tabular}{cccc|c}
    \toprule[1pt]
    \multicolumn{4}{c|}{Condensation Network} & \multicolumn{1}{c}{Evaluation Network} \\
    MiniC3D & R(2+1)D & I3D & SlowOnly & MiniC3D \\
    \hline
    \checkmark & & & & 10.1  \\
    \checkmark & \checkmark & & & 8.7   \\
    \checkmark &  & \checkmark & & 11.7  \\
    \checkmark &  &  & \checkmark & 12.1 \\
    \checkmark & & \checkmark & \checkmark & \textbf{12.4}\\
    \bottomrule[1pt]
    \end{tabular}}
    \caption{Ablation results of condensation networks on UCF101. Various combinations of condensation networks for EDC are tried.}
    \vspace{-8pt}
    \label{tab: architecture}
\end{table}

\begin{table*}[t]
    \centering
    \begin{tabular}{c|c|ccc|ccc|ccc|ccc}
    \toprule[1pt]
     \multicolumn{2}{c|}{Dataset} &  \multicolumn{3}{c|}{HMDB51} & \multicolumn{3}{c|}{UCF101} & \multicolumn{3}{c|}{SSv2} & \multicolumn{3}{c}{K400} \\
     \multicolumn{2}{c|}{IPC} & 1 & 5 & 10 & 1 & 5 & 10 & 1 & 5 & 10 & 1 & 5 & 10\\
     \multicolumn{2}{c|}{Ratio \textperthousand} & 1.2 & 5.9 & 11.8 & 0.4 & 0.9 & 4.6 & 0.18 & 0.92 & 1.8 & 0.05 & 0.23 & 0.47 \\
     \hline
     \multicolumn{2}{c|}{Full Dataset} & \multicolumn{3}{c|}{26.1} & \multicolumn{3}{c |}{52.6} & \multicolumn{3}{c|}{47.5} & \multicolumn{3}{c}{44.6} \\
     \hline
     \multirow{3}*{\makecell{Sample\\Selection}} & Random & 8.3 & 15.6 & 17.0 & 10.6 & 20.1 & 25.0 & 11.8 & 18.0 & 21.6 & 7.1 & 11.1 & 14.1 \\
     ~& Herding & 10.6 & 12.2 & 15.9 & 10.2 & 16.8 & 22.2 & 11.7 & 17.4 & 20.0 & 6.1 & 11.3 & 13.8 \\
     ~&RDED & \underline{\textbf{12.0}} & \underline{\textbf{15.7}} & \underline{\textbf{18.0}} & \underline{\textbf{14.3}} & \underline{\textbf{23.8}} & \underline{\textbf{28.2}} & \underline{12.7} & \underline{19.8} & \underline{22.5} & \underline{11.9} & \underline{16.5} & \underline{17.1} \\
     \hline
     \multirow{5}*{\makecell{Dataset\\Distillation}} &DM\textsuperscript{\textdagger}~\cite{dance} & 6.0 & 8.2 & - &-&-&-& 4.0 & 3.8 & - & 6.3 & 7.0 & - \\
     ~&MTT\textsuperscript{\textdagger}~\cite{dance} & 6.5 & 8.9 & - &-&-&-& 5.5 & 8.3 & - & 6.3 & 11.5 & - \\
     ~&FRePo\textsuperscript{\textdagger}~\cite{dance} & 8.6 & 10.3 & - &-&-&-& - & - & - & - & - & - \\
     \cline{2-14}
     ~&EDC & 6.9 & 8.9 & 12.5 & 8.6 & 9.4 & 12.0 & 8.4 & 10.5 & 12.3 & 10.9 & 11.2 & 12.6 \\
     ~&DATM & \underline{9.0} & \underline{12.9} & \underline{14.2} & \underline{14.2} & \underline{20.8} & \underline{23.2} & \underline{\textbf{12.9}} & \underline{\textbf{20.6}} & \underline{\textbf{24.2}} & \underline{\textbf{12.1}} & \underline{\textbf{16.7}} & \underline{\textbf{18.2}} \\
     \bottomrule[1pt]
    \end{tabular}
    \caption{Results of previous SOTA and our methods on HMDB51, UCF101, SSv2 and K400. Top-1 accuracy for HMDB51 and UCF101, and Top-5 accuracy for SSv2 and K400 are reported. The upper and lower sections of the table shows results for sample selection and dataset distillation, respectively. \underline{Underlined} values indicate the best results within its category (selection or distillation), while \textbf{bold} values represent the overall best result. IPC: Instance(s) Per Class, Ratio: total condensation ratio defined in ~\cref{sec:temp}. $\dag$ indicates the methods come from previous work~\cite{dance}.}
    \vspace{-8pt}
    \label{tab:sota}
\end{table*}

\subsection{Main Results and Analysis}
One of our main motivations is to compare previous image dataset condensation algorithms on video datasets. ~\Cref{fig:relabel} intuitively shows their performance on UCF101 with increasing IPCs. From the figure, we can observe that RDED with Multi-SL generally outperforms other two methods. 

\Cref{tab:sota} further shows the results on four widely-used datasets. The full dataset refers to the accuracy of the network trained on the full dataset, serving as the upper-bound results. Following the previous work~\cite{dance}, we report the top1 accuracy for HMDB51 and UCF101, and top5 accuracy for SSv2 and K400. We adopt soft labeling in evaluation for fair comparison with previous results~\cite{dance}.

For sample selection methods, our results show that Random outperforms Herding in most cases. This is violated with the intuition. However, the similar conclusion can be found in DeepCore~\cite{deepseek}. It conducts an empirical study on all kinds of traditional sample selection methods, and finds that Random remains a strong baseline. However, more recent work RDED outperforms previous sample selection methods just as results in image domains~\cite{RDED}.
For dataset distillation methods, we can find that our implementation of EDC and DATM both surpass previous methods~\cite{dance}. DATM achieves the best performance among dataset distillation methods. However, the computation and storage consumption of DATM is much higher than EDC. In our experiments, DATM consumes up to 40G GPU memories, while EDC only requires less than 10G memories. Trajectory matching methods require further refinement and development to better scale to larger datasets.

By comprehensively comparing the results across all methods, we observe that RDED achieves superior performance over DATM on HMDB51 and UCF101, while DATM performs best on the SSv2 and K400. We argue that SSv2 and K400 are more challenging than the former two datasets, primarily due to two key factors: (1) their significantly larger data scale, which demands more effective information preservation by condensation algorithms, and (2) their more diverse data distribution, characterized by a greater number of action categories and motion paradigms. 
% Specifically, SSv2 places a stronger emphasis on temporal dynamics and motion patterns, further increasing its complexity.
These challenges highlight the significant potential of dataset distillation methods in demanding scenarios. This aligns with the ultimate goal of achieving efficient training in the era of big data, where reducing computational and storage costs while maintaining performance is critical.
Last but not least, our methods achieve state-of-the-art results on all the four datasets.

\begin{table}[t]
    \centering
    \vspace{4pt}
    \begin{tabular}{c|cccc}
    \toprule[1pt]
         & \multicolumn{4}{c}{Evaluation Network} \\
         & MiniC3D & R(2+1)D & I3D & SlowOnly \\
         \hline
    Random & 14.7 & 8.8 & 9.1 & 13.7 \\
    Herding & 13.0 & 5.7 & 8.7 & 12.3 \\
    RDED & \textbf{17.7} & \textbf{11.7} & \textbf{14.1} & \textbf{21.1} \\
    \hline
    EDC & 12.0 & 3.7 & 8.7 & 7.4 \\
    DATM & 15.4 & 7.0 & 9.3 & 15.0 \\
    \bottomrule[1pt]
    \end{tabular}
    \caption{Cross-architecture generalization on UCF101 with IPC=1. MiniC3D is used as the condensation network and all the four networks are used for evaluation.}
    \vspace{-8pt}
    \label{tab:cross}
\end{table}

\noindent
\textbf{Cross-Architecture Generalization.} A key attribute of a condensed dataset is its capacity to generalize effectively to various unseen architectures~\cite{dcbench}. Previous experiments performed dataset condensation using MiniC3D and evaluated it on the same architecture. In this subsection, we evaluate the cross-architecture generalization by considering additional networks architectures for evaluation. As presented in ~\cref{tab:cross}, RDED outperforms other methods on whatever evaluation network. The rankings of different methods are consistent through different evaluation networks, and all of them shows good generalization ability.

% Our approach outperforms all methods proposed in ~\cite{dance}, achieving state-of-the-art results.

\section{Limitations}
Our results have shown that training-free interpolator fails to enhance the condensation of temporal information. However, we do not propose effective trainable interpolators to solve the problem. Besides, DATM requires relatively high resource consumption, which limits its scalability to next level. This is crucial for fully unlocking its potential in addressing challenges in the era of big data. Nevertheless, certain algorithms are not the focus of this work. Finally, there are some interesting works utilizing generative models or their ideas to assist dataset condensation~\cite{glad,minmax}. Due to the limitation of time and energy, we do not include them in our study. We anticipate future research to address these challenges, contributing to advancements in the field and enabling more efficient and effective solutions.
% One limitation of this study on video dataset condensation is that while it effectively adapts image-based condensation algorithms to video data, it may not fully capture the complexities inherent in diverse video datasets, such as variations in motion dynamics and temporal dependencies. Additional designs about motion capture and temporal processing can be incorporated. Moreover, despite achieving impressive results, the high resource consumption of DATM significantly limits its scalability to next level. Future research could focus on reducing its storage and time requirements to enhance the scalability. This is crucial for fully unlocking its potential in addressing challenges in the era of big data.
% Moreover, experiments indicate that the optimization challenges of dataset distillation remain unresolved, particularly for video data where the additional temporal dimension complicates the process. Addressing these optimization issues is essential to fully realize the potential of dataset distillation.

\section{Conclusion}
% This paper has adapted three methods representing main categories of dataset condensation methods to video domain. Our experiments reveal that, for video dataset condensation, sample diversity plays a more critical role than temporal diversity. We found that sample selection methods generally outperform existing dataset distillation approaches. However, when the condensation ratio is extremely low, dataset distillation methods can surpass sample selection methods. Finally, 

% We adapt three important dataset condensation methods from 

In response to the increasing demands of video dataset condensation, we extend three representative dataset condensation algorithms to space-time, each representing a distinct category of condensation algorithm. 
Besides, we ablate on the evaluation settings of dataset condensation and find that these settings significantly influence the final metrics, especially the labeling methods.
% Given the complexity of the evaluation phase, our experiments revealed that the evaluation setup, especially the labeling methods, has a significant impact on the final metrics. 
Therefore, we propose a unified evaluation protocol to ensure consistent and reliable assessments. Based on this evaluation protocol, we further compare temporal processing methods in videos. Our experiments reveal that our proposed sliding-window sampling outperforms previous methods. Besides, we find that the number of videos is more critical than number of frames.
% sample diversity is more critical than temporal diversity. Consequently, we introduced a co-optimized slide-window sampling method to effectively compress temporal information. 
Finally, we conducted a comprehensive comparison of various dataset condensation algorithms. Dataset distillation methods perform better in challenging scenarios, while sample selection methods excel in easier ones. This highlights the great potential of distillation methods on larger datasets. 
% While sample selection methods generally outperform distillation methods, distillation methods still maintain an advantage at extremely low compression ratios. 
We hope our study will foster future research on video dataset condensation.
\newtheorem{sketch_definition}{Definition}[section]
\definecolor{C3}{rgb}{0.839216, 0.152941, 0.156863}
% \clearpage
% \setcounter{page}{1}
% \maketitlesupplementary
\section*{Appendix}
\appendix
\section{Implementation Details of Video Condensation Algorithms}
\label{app:train}
All of the video backbones employed in this study are built upon mmaction codebase. We implement MiniC3D described in ~\cite{dance} and make use of existing R(2+1)D, I3D and SlowOnly with ResNet-18. Then, we introduce the implementation details of specific algorithms.

\subsection{Details of DATM}
We pre-compute 20 expert trajectories for each dataset. Then, we conduct a grid research for some critical hyperparameters: the mapping scope of expert trajectories, learning rate of images, syn-step (N in ~\cref{eq:datm}, M is fixed), and batch-syn on UCF101 with IPC=1. Considering the large search scope, we divide them into two groups: three values about the mapping scope as group1 and the latter three values as group2. For the hyper-parameters of other datasets, we adjust their values according to previous experience. After the hyperparameter tuning, their values are fixed across different ablations.

\subsection{Details of EDC}
We adopt the paradigm introduced in the previous subsection to perform hyperparameter tuning. The hyperparameters of EDC include batch size, learning rate, iteration and weight for different losses. Real images are used as the initialization of synthetic images both in EDC and DATM.

\subsection{Details of RDED}
Considering the concatenation operation is not employed in other methods, we simply set the factor to 1 for fair comparison with other methods. In the selection phase of RDED, we randomly sample 10 clips from each video.

\section{Evaluation Protocol}
\label{app:eval}
For fair comparison among different types of methods, we propose a unified evaluation protocol. 
We introduce the evaluation details in this section for quick reference.
% We also provide our evaluation code in the appendix, with details introduced in this section for quick reference.

\textbf{Optimizer:} we use Adam optimizer with base learning rate 0.001, betas [0.9, 0.999] and weight decay 0.01.

\textbf{Scheduler:} we train the evaluation for 300 epochs with a cosine learning rate scheduler. 

\textbf{Batch size:} following the implementation in SelMatch~\cite{selmatch}, we adjust batch size according to dataset and IPC to make sure the iteration number is fixed:

\begin{equation}
    \text{BS} = \text{base} \times \text{IPC},
\end{equation}

\noindent
where BS is the training batch size, and base is the batch size when IPC is equal to one. The base for HMDB51 and UCF101 is set to 10, base for SSv2 is 20, and that for K400 is 40.
% (~\cref{tab:bs}). 

\textbf{Augmentation:} we apply resized crop and horizontal flip when training evaluation models.

\textbf{Loss Function:} we use MSE-GT as the loss function.

\textbf{Labeling Method:} we use soft labeling in \cref{tab:sota} for comparison with previous results, and use Multi-SL in other experiments.

\section{More Results}
\label{app:exp}
~\Cref{supp:inter} supplements ablation results of temporal processing on SSv2. It shares similar observations as previous results: sliding-window sampling outperforms previous segment sampling, and training-free interpolation brings sharp drops in performance.

~\Cref{supp:sota} supplements results on UCF101 with Multi-SL. The results are higher than those in ~\cref{tab:sota} with soft labeling. The comparison between methods is consistent with the previous observation: RDED and DATM achieve best results within their respective category, and RDED achieves sota on UCF101.

\section{Visualization}
\begin{figure*}
    \centering
    \includegraphics[width=0.87\linewidth]{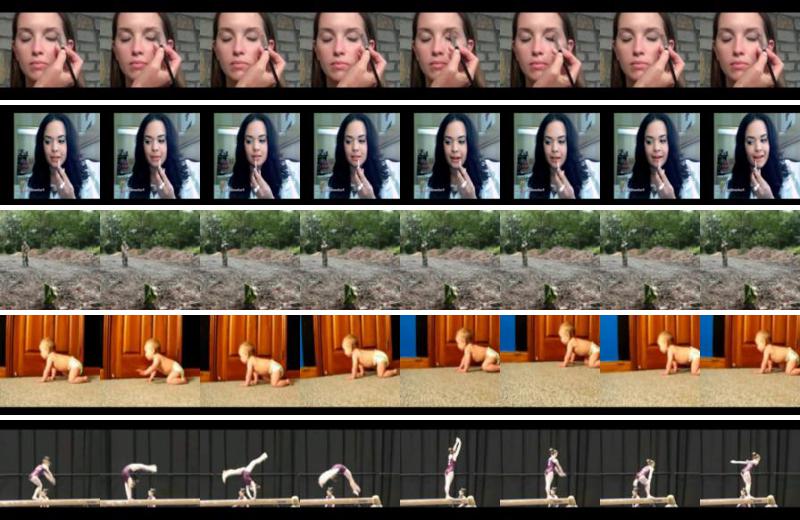}
    \caption{\textbf{Visualization of RDED for UCF101 IPC=1}}
    \label{fig:vis-rded}
\end{figure*}

\begin{figure*}
    \centering
    \includegraphics[width=0.87\linewidth]{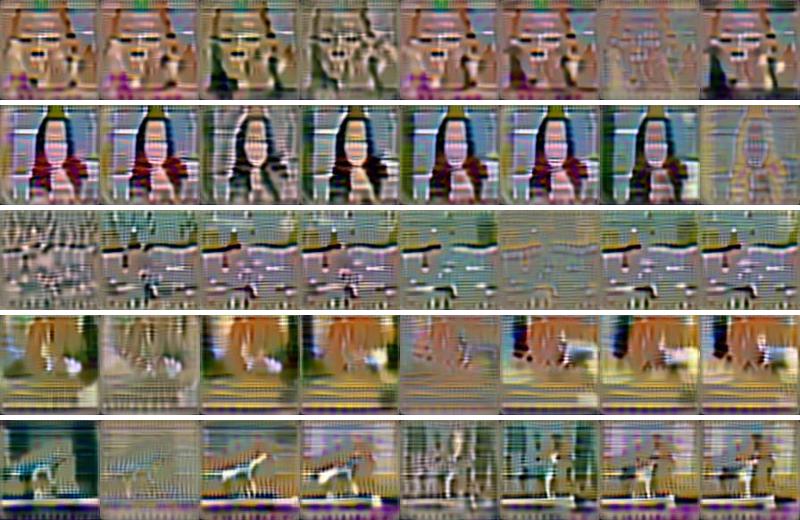}
    \caption{\textbf{Visualization of EDC for UCF101 IPC=1}}
    \label{fig:vis-edc}
\end{figure*}

We show the visualization of RDED and EDC in ~\cref{fig:vis-rded} and ~\cref{fig:vis-edc}, representing selection and distillation method respectively. Compared with RDED (real data), EDC captures some crucial patterns through distillation. While there is some redundant information in the temporal dimension.

\begin{table}[t]
    \centering
    \setlength{\tabcolsep}{4pt}
    \begin{tabular}{lc|cc}
    \toprule
       Sampling  & Interopolation & RDED & EDC \\
         \hline 
        None& \ding{55} & 17.5 & 11.2 \\
        Segment & \ding{55} & 18.8 & 11.3  \\
        Sliding-window & \ding{55} & \textbf{21.1} & \textbf{13.1}   \\
        Sliding-window & \checkmark & 13.4 & 12.6  \\
        \bottomrule
    \end{tabular}
    \caption{Ablation results of temporal processing on SSv2. The setting is same as ~\cref{tab:inter}. It also shows that sliding-window sampling outperforms segment sampling. Linear interpolation does not work for all methods.}
    \label{supp:inter}
\end{table}

\begin{table}[t]
    \centering
    \begin{tabular}{c|c|ccc}
         \toprule
         \multicolumn{2}{c|}{IPC} & 1 & 5 & 10 \\
         \multicolumn{2}{c|}{Full Dataset} & \multicolumn{3}{c}{52.6} \\
         \hline
         \multirow{3}*{\makecell{Sample\\Selection}} & Random & 14.7 & 26.9 & 31.4 \\
         ~ & Herding & 13.0 & 22.8 & 27.7 \\
         ~ & RDED & \underline{\textbf{17.7}} & \underline{\textbf{29.4}} & \underline{\textbf{33.5}} \\
         \hline
         \multirow{2}*{\makecell{Dataset\\Distillation}} & EDC & 12.0 & 22.2 & 24.3 \\
         ~ & DATM & \underline{15.4} & \underline{27.2} & \underline{31.7} \\
         \bottomrule
    \end{tabular}
    \caption{Results of our methods with Multi-SL on UCF101.}
    \label{supp:sota}
\end{table}

\section*{Acknowledge}
Yang Chen would like to thank Ruopeng Gao, Shuai Wang and Shitong Shao for the kind discussion.

\newpage

{
    \small
    \bibliographystyle{ieeenat_fullname}
    \bibliography{main}
}

% \newpage
% \input{sec/X_suppl.tex}

\end{document}